\documentclass[10pt,twocolumn,letterpaper]{article}

\usepackage[pagenumbers]{wacv} 
\usepackage{pifont}

\definecolor{wacvblue}{rgb}{0.21,0.49,0.74}
\usepackage[pagebackref,breaklinks,colorlinks,allcolors=wacvblue]{hyperref}

\usepackage[accsupp]{axessibility} 

\usepackage{multirow}
\usepackage{diagbox}

\usepackage{caption}
\def\wacvPaperID{2875} 
\def\confName{WACV}
\def\confYear{2026}

\title{Overcoming Small Data Limitations in \\ Video-Based Infant Respiration Estimation}

\author{Liyang Song$^\dagger$, Hardik Bishnoi$^\dagger$, Sai Kumar Reddy Manne,\\ Sarah Ostadabbas, Briana J. Taylor, Michael Wan$^\ddagger$\\\\
Northeastern University\\
$^\dagger$\small\textit{equal contribution} \quad
$^\ddagger${\tt\small mi.wan@northeastern.edu}\\
}

\begin{document}
\maketitle
\begin{abstract}
The development of contactless respiration monitoring for infants could enable advances in the early detection and treatment of breathing irregularities, which are associated with neurodevelopmental impairments and conditions like sudden infant death syndrome (SIDS). But while respiration estimation for adults is supported by a robust ecosystem of computer vision algorithms and video datasets, only one small public video dataset with annotated respiration data for infant subjects exists, and there are no reproducible algorithms which are effective for infants. We introduce the \textbf{a}nnotated \textbf{i}nfant \textbf{r}espiration dataset of \textbf{400} videos (AIR-400), contributing 275 new, carefully annotated videos from 10 recruited subjects to the public corpus. We develop the first reproducible pipelines for infant respiration estimation, based on infant-specific region-of-interest detection and spatiotemporal neural processing enhanced by optical flow inputs. We establish, through comprehensive experiments, the first reproducible benchmarks for the state-of-the-art in vision-based infant respiration estimation. We make our dataset, code repository, and trained models available for public use. 
\end{abstract}
    
\section{Introduction}
\label{sec:intro}

\begin{figure*}[!h]
    \centering
    \includegraphics[width=\linewidth]{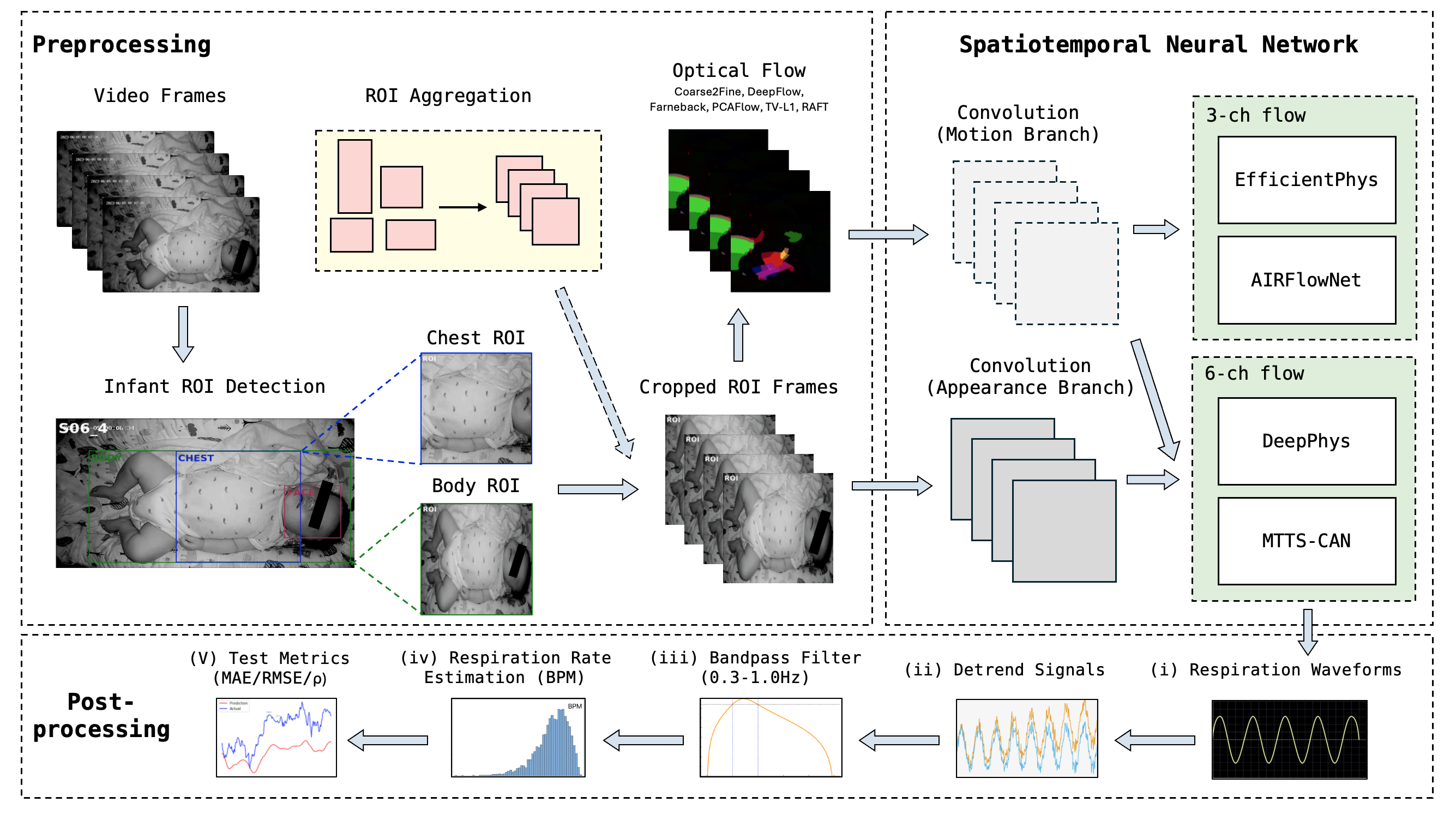}
    \vspace{-8mm}
    \caption{\textbf{Our infant respiration estimation pipeline}  combines a preprocessing module with infant-specific spatial region-of-interest (ROI) detectors, spatiotemporal inference from neural networks, and signals post-processing. We implement two region-of-interest (ROI) strategies, an infant \textbf{body ROI} detector, and infant \textbf{chest ROI} detector. Clips are cropped to the ROI and processed by optical flow (Coarse2Fine \cite{liu2009beyond}, DeepFlow \cite{weinzaepfel2013deepflow}, Farneb\"ack \cite{farneback2003two}, PCAFlow \cite{wulff2015efficient}, TV-L1 \cite{zach2007aduality, sanchez2013tvl1}, RAFT \cite{zachary2020raft}). Spatiotemporal neural networks (DeepPhys \cite{chen2018deepphys}, MTTS-CAN \cite{liu2020multi}, EfficientPhys \cite{liu2023efficientphys}, AIRFlowNet \cite{manne_automatic_2023}) then process appearance and motion inputs, and post-processing via detrending and bandpass filtering is applied to obtain respiration rates and waveforms.}
    \label{fig:architecture}
\end{figure*}

In infancy, healthy respiration is critical for optimal brain development and early detection and treatment of irregular respiration is critical to prevent neurodevelopmental impairment \cite{janvier2004apnea}. Breathing irregularities such as apnea or respiratory distress are significantly more common in high-risk neonates who often require post-delivery monitoring in neonatal intensive care units (NICUs) \cite{reuter2014respiratory, eichenwald2016apnea}. However, respiration monitoring the NICU relies on contact sensors which can damage and cause pain to the sensitive skin of a neonate \cite{bonner2017there}. Moreover, gold standard respiration monitoring cannot be easily conducted in the home, post-discharge, despite substantial evidence linking apnea and sudden infant death syndrome (SIDS) \cite{martin2022apnea}. The rise of human- and health-centric computer vision algorithms and widespread adoption of computationally-enabled baby monitors presents the opportunity to develop a contactless alternative to infant respiratory detection in home and medical settings. Recent advances in computer vision have included deep learning algorithms designed and trained to track infant-specific poses, states, and actions \cite{huang2021invariant,wan_infanface_2022,huang_posture-based_2023,dechemi_ebabynet_2024}. Progress in infant respiration waveform and rate estimation, however, has been limited by the scarcity of annotated video data \cite{ostadabbas2015vision,ostadabbas2014passive}. As we will explain, most existing datasets are private and cannot be easily used or validated, and the sole work featuring a public dataset and algorithm for this task faces major limitations due to issues with dataset size and model reproducibility.

In this paper, we establish infant respiration estimation as a reproducible and benchmarkable task in computer vision for the first time, through three key contributions. First, we introduce a new video dataset with carefully synchronized and annotated respiration waveforms, which triples the amount of publicly accessible data, in terms of both the number of clips and the diversity of recruited infant subjects. Second, we propose a novel method for infant respiration waveform estimation that leverages infant pose detection and fine-grained region-of-interest (ROI) tracking, as illustrated in \Cref{fig:architecture}. Third, we perform the most comprehensive evaluation to date of infant respiration estimation pipelines, systematically comparing optical flow and spatiotemporal network architectures, and rigorously testing reproducibility and generalizability across experimental settings. Together, these contributions provide a reliable foundation for advancing infant respiration monitoring within the computer vision community. Our dataset, code repository, and trained models can be found at \url{https://github.com/michaelwwan/air-400}.

\section{Related Work}
\label{sec:related}

Video-based physiological waveform and rate estimation has advanced rapidly. One of the most successful early models, DeepPhys \cite{chen2018deepphys}, estimated both the cardiovascular and respiratory amplitude waveform by tracking changes in skin reflectance using a convolutional neural network with an attention mechanism to analyze differences in nearby frames. Subsequent methods found success applying 3D convolutional neural networks \cite{yu_remote_2019}, though at a higher computational cost. Newer models such as the multi-task temporal shift convolutional attention network
(MTTS-CAN) \cite{liu2020multi} and EfficientPhys \cite{liu2023efficientphys} reduce the computational burden by replacing 3D convolutional networks with temporal shift modules combined with 2D convolutional processing. 

Alongside methodological advances, vision-based physiological estimation has benefited from the public release of video datasets featuring a large number of adult subjects, accompanied by carefully calibrated ground truth physiological signals. Most notably, these include the COHFACE dataset \cite{heusch2017reproducible}, the MAHNOB dataset \cite{soleymani2011multimodal}, and the synthetic SCAMPS \cite{mcduff2022scamps} dataset. Characteristics of these datasets are tabulated alongside public and private datasets featuring both adult and infant subjects in \Cref{tab:dataset-table}. Focusing on respiration in the adult domain, there are 687 ground-truth videos from 67 real subjects publicly available with ground truth respiration signals, in addition to 2,800 videos from 2,800 synthetic subjects from SCAMPS. 

By contrast, infant respiration estimation has been constrained by severe data scarcity, due to the complexity of collecting infant video data with corresponding ground truth respiration waveforms. The only public, annotated video data on infant respiration comes from our previous work in Manne et al. \cite{manne_automatic_2023}, which introduced the \textbf{a}nnotated \textbf{i}nfant \textbf{r}espiration dataset of \textbf{125} videos, or \textbf{AIR-125}, consisting of 125 short video clips drawn from eight infant subjects. This provided just enough data to train and test a preliminary infant respiration estimation model. Unfortunately, our more recent and comprehensive analysis documented in \Cref{sec:results} shows that the strong numerical results obtained in \cite{manne_automatic_2023} do not generalize to other choices of train--test split in the AIR-125 dataset, and are also not reproducible in exhaustive cross-validation testing in our newly expanded dataset. In this paper, we aim to rectify issues of data scarcity and model reproducibility with our updated dataset, methodology, and experimental contributions. 

\begin{table}
    \centering
        \caption{\textbf{Datasets available to support video-based human respiration estimation}, highlighting the extreme scarcity of video data available featuring infant subjects. PPG: photoplethysmogram, PR: pulse rate, RR: respiratory rate, Resp: respiration waveform, AU: action unit, BVP: blood volume pulse, EEG: electroencephalogram, ECG: electrocardiogram, SpO2: blood oxygenation.}
    \vspace{-3mm}
    \resizebox{\linewidth}{!}{
    \begin{tabular}{lllrrc}
        \toprule
         \textbf{\textsc{Dataset}} & \textbf{\textsc{Ground Truth}} & \textbf{\textsc{Domain}} & \textbf{\textsc{Videos}} & \textbf{\textsc{Subjects}} &  \textbf{\textsc{Public}}  \\
         \midrule
         SCAMPS \cite{mcduff2022scamps} & PPG, PR, RR, Resp, AU & Adult & 2,800 & 2,800 & \ding{51} \\ 
         COHFACE \cite{heusch2017reproducible} & Resp, BVP & Adult & 160 & 40 & \ding{51}
         \\
         MAHNOB \cite{soleymani2011multimodal} & ECG, EEG, Resp & Adult & 527 & 27 & \ding{51} 
         \\
         AFRL \cite{estepp2014recovering} & ECG, EEG, PPG, PR, RR & Adult & 300 & 25 & \ding{55}  \\
         OBF \cite{li2018obf} & RR, PPG, ECG & Adult & 212 & 106 & \ding{55}  \\
         \midrule
         Villarroel \textit{et al.} \cite{villarroel_non-contact_2019} & Resp, PPG, SpO2 & Infant & 384 & 30 & \ding{55} \\
         Földesy \textit{et al.} \cite{foldesy2020reference} & Resp & Infant & 1,440 & 7 & \ding{55} \\
         Kyrollos \textit{et al.} \cite{kyrollos2021noncontact} & Resp & Infant & 20 & 1 & \ding{55} \\
         Lorato \textit{et al.} \cite{lorato2021towards} & Resp & Infant & 90 & 2 & \ding{55} \\
         Tveit \textit{et al.} \cite{tveit2016motion} & RR & Infant & 6 & 2 & \ding{55} \\
         AIR-125 \cite{manne_automatic_2023} & Resp, RR, Pose & Infant & 125 & 8 & \ding{51} \\
         \midrule
         \textbf{AIR-400 \textit{(Ours)}} & Resp, RR, Pose & Infant & 400 & 18 & \ding{51} \\
         \bottomrule
    \end{tabular}
    }
    \label{tab:dataset-table}
\end{table}

\section{An Expanded Infant Respiration Dataset}
\label{sec:dataset}

\begin{figure}[!ht]
    \centering
    \includegraphics[width=\linewidth]{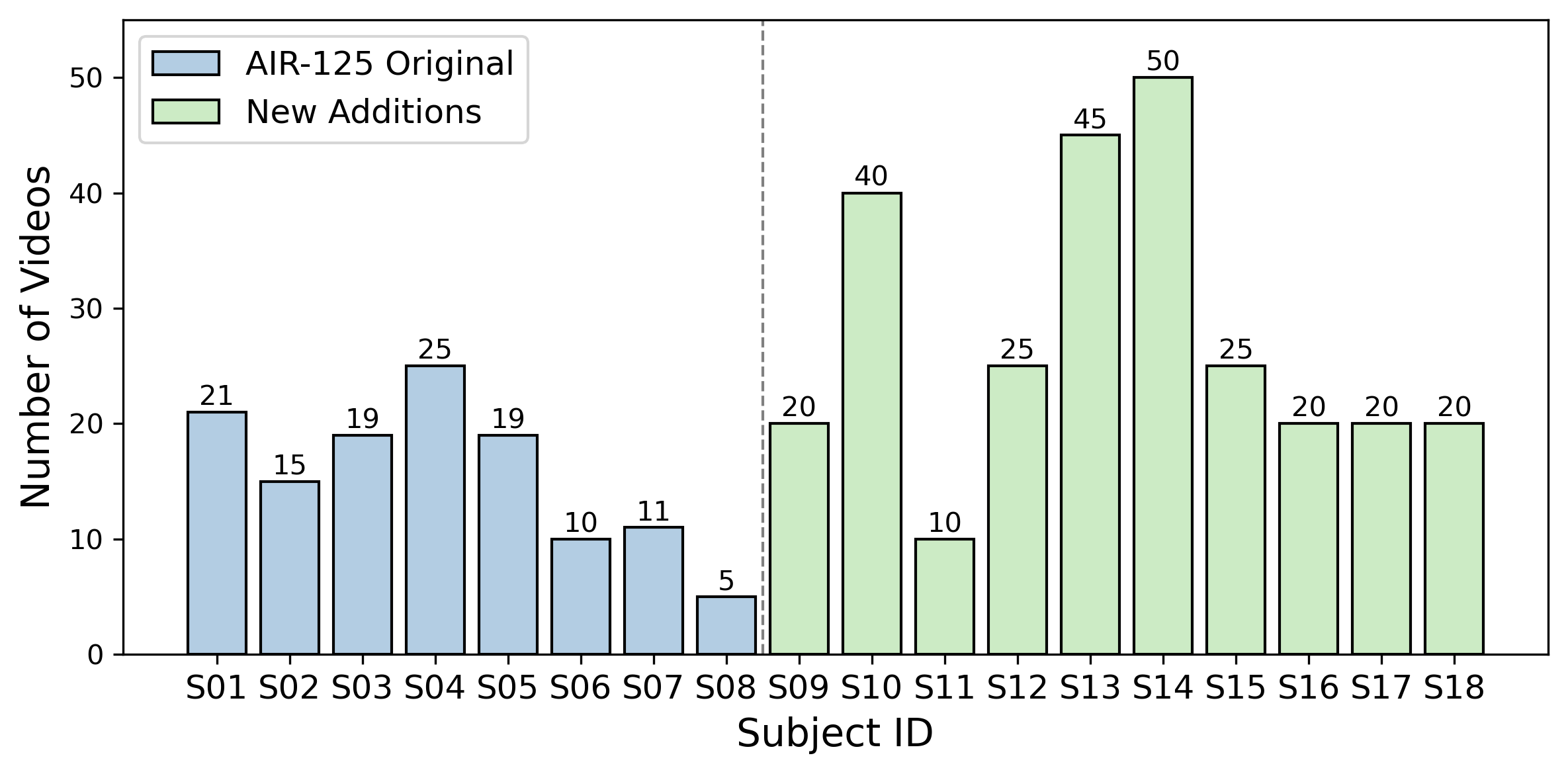}
    \vspace{-8mm}
    \caption{\textbf{Number of videos per subject in AIR-400.} The dataset comprises 400 videos from 18 infant subjects. Blue bars (S01--S08) represent the AIR-125 dataset \cite{manne_automatic_2023} (125 videos from 8 subjects), while green bars (S09--S18) show our additions (275 videos from 10 new subjects from the same study as in \cite{manne_automatic_2023}). Each bar indicates the number of 60\,s clips available for that subject.}
    \label{fig:subject_breakdown}
\end{figure}

We expand our previous dataset of 125 infant respiration videos from Manne et al. \cite{manne_automatic_2023} with an additional 275 clips drawn from new subjects from the same study population, formatted and annotated by our interdisciplinary team in the same way. We package the datasets together to produce a unified \textbf{a}nnotated \textbf{i}nfant \textbf{r}espiration dataset of \textbf{400} videos, or \textbf{AIR-400}. We next describe our process for collecting, curating, and annotating the 275 new videos, and characterize the AIR-400 dataset as a whole. See \Cref{fig:subject_breakdown} for a breakdown of AIR-125 and AIR-400 by subject. 

\subsection{Data Collection and Curation}

\label{sec:curation}

\begin{figure*}[!h]
    \centering
    \includegraphics[width=\linewidth]{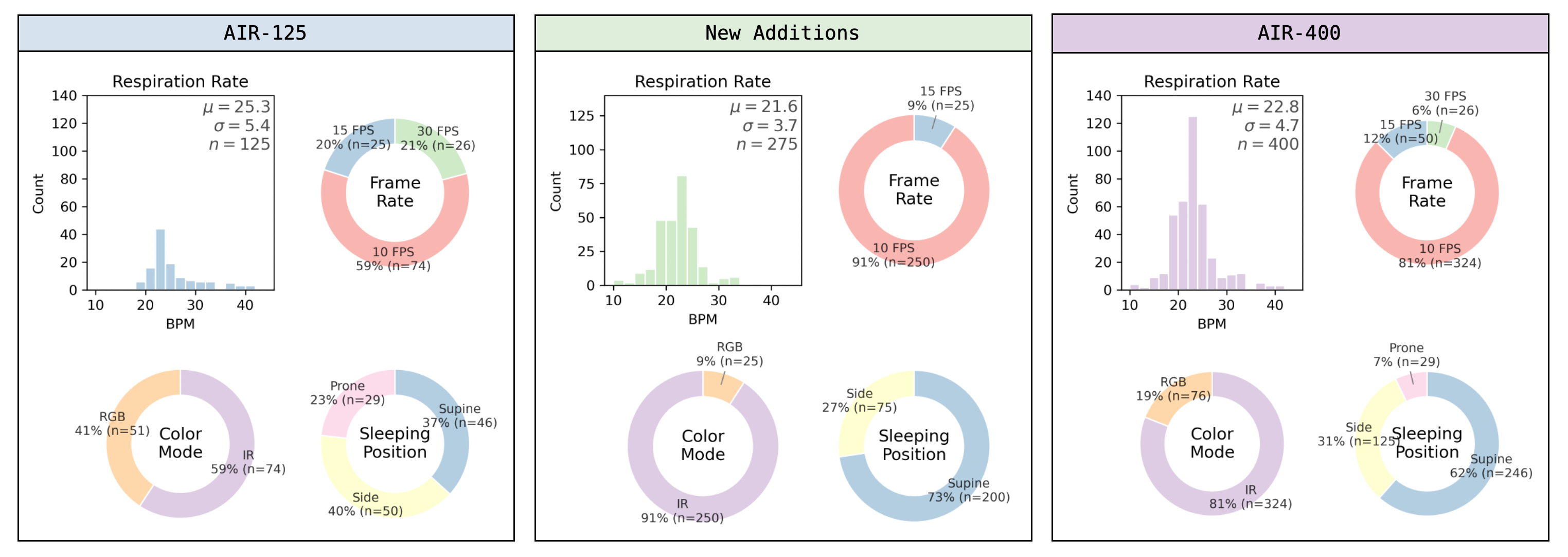}
    \vspace{-7mm}
    \caption{\textbf{Dataset statistics for AIR-125 \cite{manne_automatic_2023}, our newly contributed 275 videos, and the combined AIR-400.} Each panel shows a respiration-rate histogram, plus ring charts for frame rate, color mode (RGB vs infrared (IR)), and sleeping position (supine, side, and prone). The expanded dataset provides broader respiratory pattern diversity and varied recording conditions.}
    \label{fig:dataset_characteristics}
\end{figure*}

We collected ${\sim}100$ hours of overnight baby monitor footage from 10 more infant subjects from the same study as in \cite{manne_automatic_2023}, under Northeastern Institutional Review Board approval (IRB \#22-11-32). Participants ranged from 4--11 months in age. We provided baby monitor cameras to caregivers, with instructions to mount them overhead for a whole-body view of the infant. Raw recordings were captured for hours-long sessions overnight, and infant sleep is punctuated by periods of arousal and waking, as well as caretaker interactions. Footage was captured in both RGB and infrared (IR) modes, automatically determined by the camera based on the availability of ambient lighting.

Technical constraints presently stand in the way of fully-automated pediatric health monitoring from long-form video footage. Extreme infant movements and postures and anomalous events such as caretaker interactions can hamper effective tracking. High computational costs of preprocessing and training also prevent rapid experimentation needed at the research stage. Thus, in order to provide a feasible starting point for infant respiration estimation, we follow AIR-125 and curate a set of 60\,s videos, featuring relatively clean samples of infant respiration behavior.

We were guided by three primary criteria to ensure dataset quality and utility. First, we favored highly visible chest and abdomen movements to enable effective visual annotation of respiratory cycles. We avoided videos with obstructions such as blankets, pacifiers, or toys. Second, we excluded excessive subject movement, caretaker interactions, or camera shake, which obscure respiratory movements. This consideration was especially important for the IR footage, where the combination of subtle thoracic movements, lower frame rates, and increased temporal noise in underexposed environments can create ambiguity. Third, we favored supine (face-up) positions for optimal chest visibility, while also including other sleeping orientations for variety. We also ensured diversity in respiratory patterns by selecting videos across an extended range of infant breathing rates (15--40 breaths-per-minute (BPM)). This range encompasses both typical clinical norms (30--60\,BPM) \cite{fleming_normal_2011} and challenging real-world scenarios where measured rates appear lower due to movement artifacts or natural breathing pauses during infant sleep. See \Cref{fig:dataset_characteristics} for an overview of composition of the resulting dataset.

\subsection{Annotation Process}

To derive the ground truth respiration waveforms, we used the VGG Image Annotator software \cite{dutta2019vgg} to systematically analyze respiratory cycles and manually mark the peaks of the respiration waveform throughout each video sequence. The annotation was performed by one of the authors through iterative video playback, with frequent pausing, frame-by-frame advancement, and backtracking to ensure accurate peak identification. Respiration peaks were defined as the exhalation onset point (the transition from inspiration to expiration) \cite{massaroni2024advances}, identified by observing the reversal from chest expansion to contraction. In cases where this transition was ambiguous, the minimum chest or abdomen position (point of maximum compression) was used as an alternative marker \cite{santos2023realtime,kohn2015monitoring}.

Identification of peaks relied on multiple visual indicators: direct thoracic and abdominal movements, facial cues (nostril and mouth changes), indirect movements propagated to other body regions (shoulders, limbs), and clothing deformation patterns which often amplified subtle respiratory motion. This multi-marker approach was essential given the variability in infant positioning and partial occlusions common in naturalistic recordings. Each video received at least two complete annotation passes: an initial marking of respiration peaks followed by validation for temporal consistency and physiological plausibility. 

Following \cite{manne_automatic_2023}, we converted discrete peak annotations into continuous respiration waveforms, as follows. First, an impulse signal $I(t)$ was generated with unit impulses at annotated peak timestamps:
\begin{equation} 
I(t) = \sum_{i} \delta(t - t_i),
\label{eq:annotation-to-signal}
\end{equation}
where $t_i$ is the timestamp of the $i$th annotated respiration peak. This impulse signal was then convolved with a Gaussian kernel to produce a smooth waveform:
\begin{equation} 
w(t) = (I * G_\sigma)(t) = \sum_{i} G_\sigma(t - t_i),
\label{eq:signal-to-waveform}
\end{equation}
where $G_\sigma$ is a Gaussian kernel with standard deviation $\sigma$. We choose $\sigma = 4$ empirically to produce physiologically plausible waveforms while accommodating temporal uncertainties inherent in manual annotation. 

\subsection{Dataset Statistics}

The AIR-400 dataset adds 275 additional annotated infant respiration videos to an existing public corpus of 125, providing 400 annotated clips in all, from 18 infant subjects, as illustrated in \Cref{fig:subject_breakdown}. \Cref{fig:dataset_characteristics} presents the distributional characteristics of both components. The respiration rate distributions show complementary coverage, with AIR-125 exhibiting a mean of 25.3\,BPM ($\sigma = 5.4$) and the new additions centering at 21.6\,BPM ($\sigma = 3.7$). This combined distribution spans a broader physiological range (approximately 15--40\,BPM), as discussed in Section 3.2, capturing both normal tidal breathing and elevated respiratory patterns common in infant monitoring scenarios. 

Notable differences in recording modalities reflect technical constraints inherent to infant monitoring. The predominance of infrared footage in the new additions (91.5\% IR vs 8.5\% RGB) corresponds to overnight recording conditions where ambient lighting necessitates IR capture. This modality difference directly influences frame rate characteristics: IR recordings are constrained to 10 frames per second (fps) due to sensor limitations, while RGB captures operate at higher frame rates (15--30\,fps). The original AIR-125 dataset's mixed frame rate distribution (59.2\% at 10\,fps, 20.8\% at 30\,fps) reflects its more balanced RGB--IR composition (40.8\% RGB, 59.2\% IR). This diversity in temporal resolution enables researchers to evaluate algorithm robustness across the heterogeneous capture conditions encountered in real-world deployment, where consumer baby monitors operate at varying frame rates depending on lighting conditions.

Sleeping position distributions differ between the collections, with the new additions containing no prone position recordings. This reflects a methodological choice in our curation criteria (\Cref{sec:curation}), as prone positions occlude thoracic and abdominal movement markers, compromising annotation reliability. The respiratory signal in prone infants manifests primarily through subtle spinal and shoulder blade movements, which were challenging to annotate consistently compared to the clear thoracoabdominal excursions visible in supine (74.6\%) and side (25.4\%) positions. While the inclusion of prone positions (23.2\%) in AIR-125, alongside supine (36.8\%) and side (40\%) positions, increases variety at the cost of annotation reliability. We opted to focus on reliability, in line with our goal of establishing the first robust benchmark and reproducible method for the infant respiration estimation task. 

\section{Methodology}
\label{sec:methodology}

Our infant respiration estimation pipeline, illustrated in \Cref{fig:architecture}, combines a preprocessing module with infant-specific spatial region-of-interest (ROI) detection with spatial neural network processing. We work in particular with an infant-specific body ROI detector, together with a novel face-aware chest ROI detector, designed to exploit the pose and camera geometry of baby monitors to isolate the infant chest area. The crop from our infant ROI detector is then enhanced with deep, fine-grained optical flow, and processed by a spatiotemporal neural network optimized to track physiological signals. Post-processing is applied to convert the output respiration waveform into a respiration rate estimation for evaluation. 

\subsection{Preprocessing: ROI and Optical Flow}
\label{sec:preprocessing}

\begin{figure}[!ht]
    \centering
    \includegraphics[width=\linewidth]{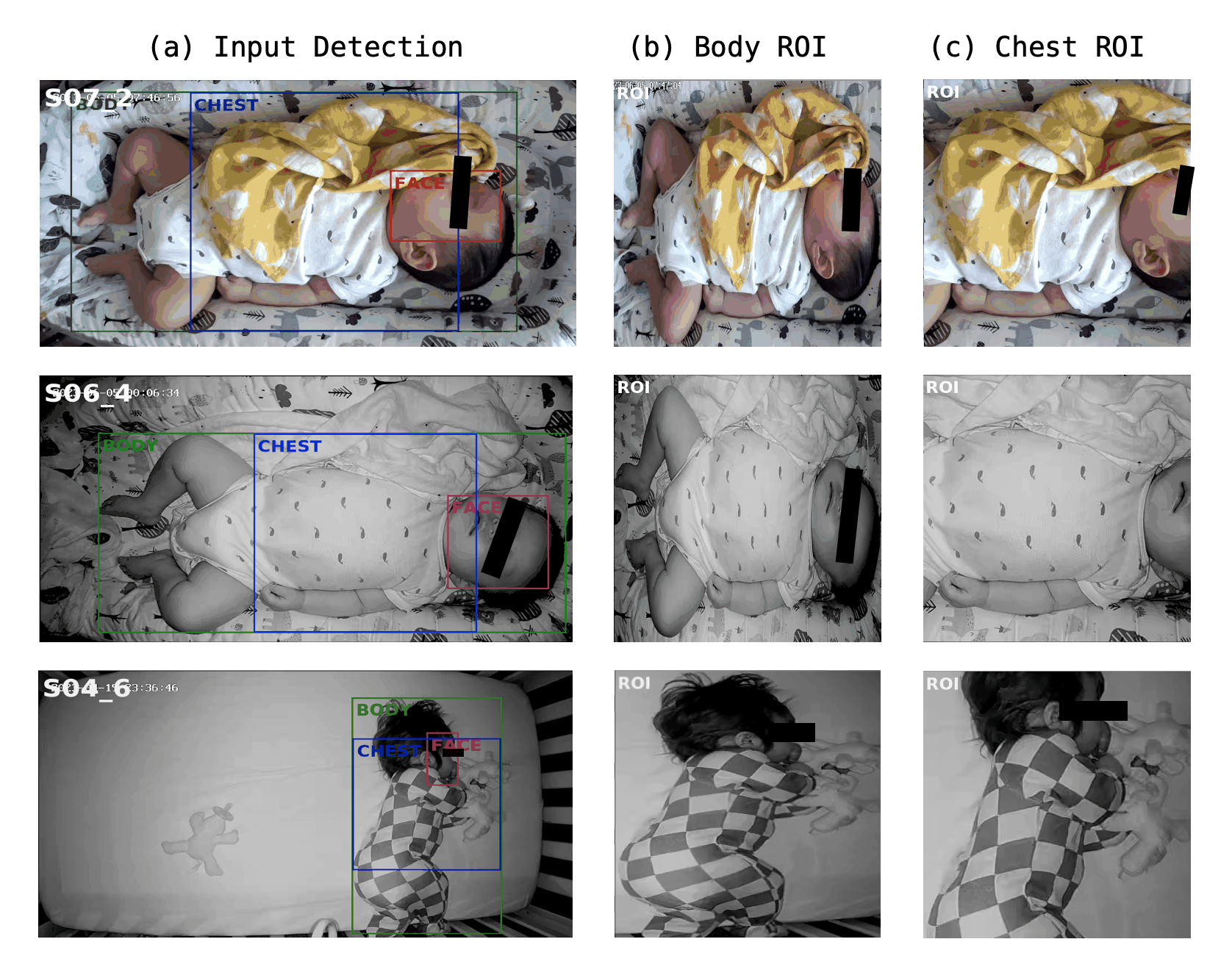}
    \vspace{-8mm}
    \caption{\textbf{Infant body region-of-interest (ROI) and chest ROI detections} across three subjects.}
    \label{fig:roi_boxes}
\end{figure}

Visual evidence of the respiratory process is most prominent in the chest area, but existing techniques for identifying the chest ROI fall short in the infant domain, primarily due to poor performance of face detection for infant subjects in-crib. We test two variations on an infant-specific method for identifying and cropping to the relevant area in the video feed, \textbf{body ROI} detector and \textbf{chest ROI} detector, defined as follows, and illustrated in \Cref{fig:roi_boxes}.

\begin{figure}[!ht]
    \centering
    \includegraphics[width=\linewidth]{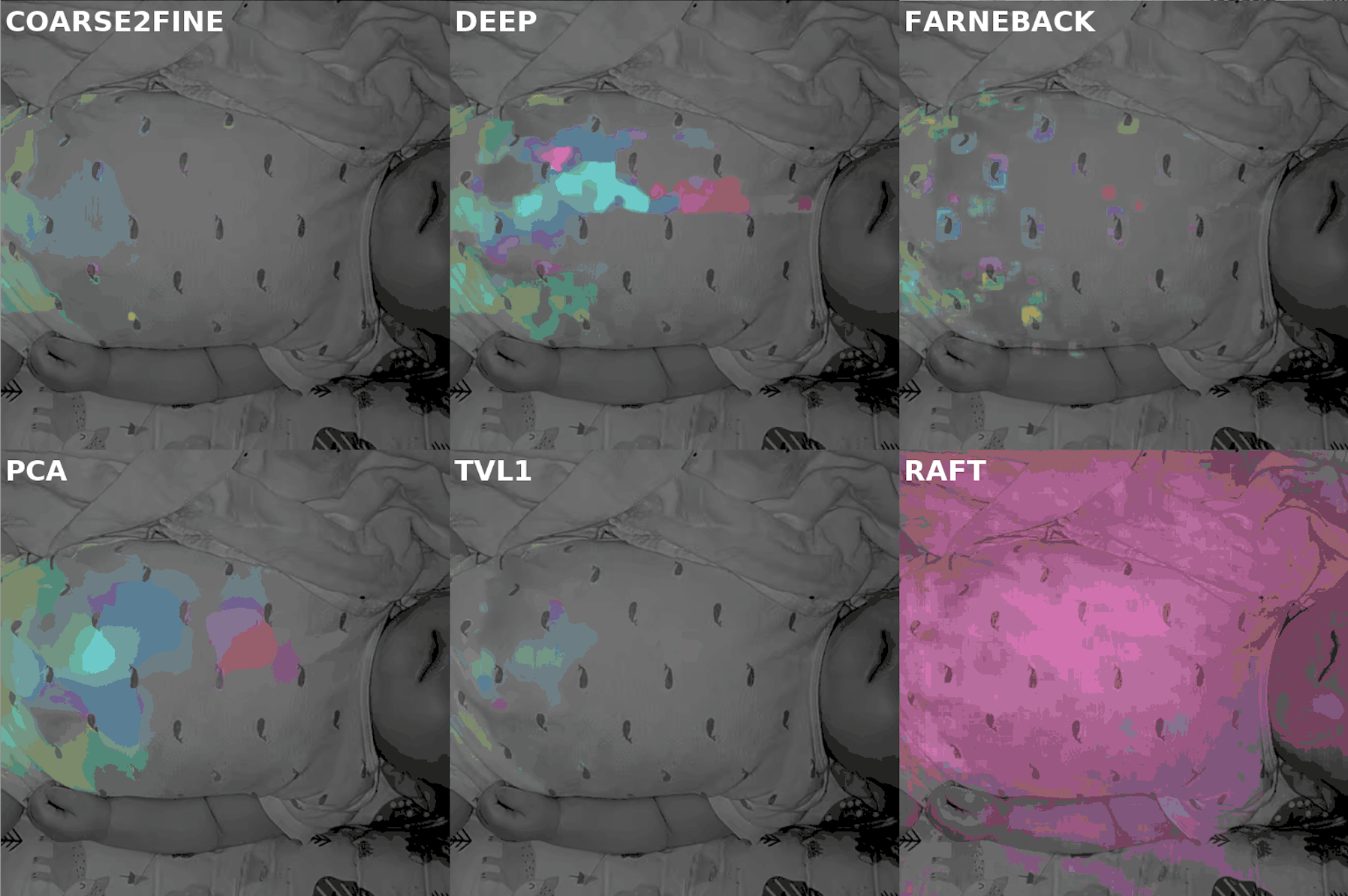}
    \vspace{-6mm}
    \caption{\textbf{Visualization of optical flow estimation on a sample frame from AIR-400.} Motion fields produced by six optical flow algorithms: Coarse2Fine, DeepFlow, Farneb\"ack, PCAFlow, TV-L1, RAFT.}
    \label{fig:optical_flow_example}
\end{figure}

For each video frame $X_t \in \mathbb{R}^{H\times W\times 3}$, we apply YOLOv8 \cite{Jocher_Ultralytics_YOLO_2023} to obtain the infant body bounding box $B_t=[x^B_t,y^B_t,w^B_t,h^B_t]$, and YOLOv8-Face \cite{Jocher_Ultralytics_YOLO_2023}, which is a lightweight model finetuned on human faces, to detect the infant face $F_t=[x^F_t,y^F_t,w^F_t,h^F_t]$. To mitigate infant face detection failures due to pose angle variation \cite{wan_infanface_2022}, face detection is applied to $X_t$ and its $90^\circ/180^\circ/270^\circ$ rotations, and the most confident prediction selected. Detections are performed every $s$ frames (default $s=10$) and at the last frame. Let $\mathcal{T}$ denote the set of detection indices. (Trivial detections are dropped for subsequent aggregation, if none remain, we fall back to the full frame (for body ROI) or a centered square of side $\min(W,H)$ (for chest ROI). All crops are clamped to image bounds, with out-of-bounds regions zero-padded.)

\textbf{Body ROI.} If a valid person detection is present, the per-frame body ROI is $B_t$. If no person is detected (or a degenerate full-frame box is returned), we mark $B_t$ as unreliable and fall back to the full frame for that time step. From the set $\{B_t\}_{t\in\mathcal{T}}$, we compute a single representative body ROI,
\begin{equation} 
\tilde{B} = [\tilde{x}^B,\tilde{y}^B,\tilde{w}^B,\tilde{h}^B],
\label{eq:body-roi}
\end{equation}
by aggregating centers with the median and sizes with a robust upper quantile:
\begin{equation}
\tilde{c}^B_x=\operatorname{median}_t\!\left(x^B_t+\tfrac{1}{2}w^B_t\right)
\label{eq:body-roi-cx}
\end{equation}
\begin{equation}
\tilde{c}^B_y=\operatorname{median}_t\!\left(y^B_t+\tfrac{1}{2}h^B_t\right)
\label{eq:body-roi-cy}
\end{equation}
\begin{equation}
\tilde{w}^B=\operatorname{P75}_t(w^B_t)
\label{eq:body-roi-w}
\end{equation}
\begin{equation}
\tilde{h}^B=\operatorname{P75}_t(h^B_t),
\label{eq:body-roi-h}
\end{equation}
and then recentering to $(\tilde{c}^B_x,\tilde{c}^B_y)$ and clamping to the image bounds. Optionally, we enlarge $\tilde{B}$ by a factor $k{>}1$ and re-clamp.

\textbf{Chest ROI.} Given $B_t$ and $F_t$, define their centers
$b_t=\big(x^B_t+\tfrac{1}{2}w^B_t,\; y^B_t+\tfrac{1}{2}h^B_t\big)$ and
$f_t=\big(x^F_t+\tfrac{1}{2}w^F_t,\; y^F_t+\tfrac{1}{2}h^F_t\big)$.
We construct a face-aware chest center $c^C_t$ by nudging the body center toward the face by a factor $\alpha\in(0,1)$, but only along the body’s longer axis to preserve anatomical plausibility:
\begin{equation} 
c^C_t=
\begin{cases}
\big(b_{x,t} + \alpha(f_{x,t}-b_{x,t}),\; b_{y,t}\big) & \text{if } w^B_t \ge h^B_t,\\[4pt]
\big(b_{x,t},\; b_{y,t} + \alpha(f_{y,t}-b_{y,t})\big) & \text{otherwise.}
\end{cases}
\label{eq:chest-roi-center}
\end{equation}
The chest box side length is the body’s short side,
$s_t=\min(w^B_t,h^B_t)$, yielding the per-frame square
$C_t=[x^C_t,y^C_t,s_t,s_t]$ centered at $c^C_t$ and constrained to lie inside $B_t$ and the image. If $F_t$ is missing, we substitute $f_t\leftarrow b_t$ (no nudge). Aggregation mirrors the body ROI: median center, 75th-percentile side, recentering, clamp, and optional enlargement, producing $\tilde{C}=[\tilde{x}^C,\tilde{y}^C,\tilde{s},\tilde{s}]$.

\textbf{Optical Flow.} We compute dense motion within the crop using one of: Coarse2Fine \cite{liu2009beyond}, DeepFlow \cite{weinzaepfel2013deepflow}, Farneb\"ack \cite{farneback2003two}, PCAFlow \cite{wulff2015efficient}, TV-L1 \cite{zach2007aduality, sanchez2013tvl1}, or RAFT \cite{zachary2020raft}, all illustrated in \Cref{fig:optical_flow_example}. If desired, frames are resampled to a target rate $f_s'$ before flow. The resulting flow $U\in\mathbb{R}^{T\times h\times w\times 3}$ (channels: $u$, $v$, magnitude) is normalized alongside the frames; if flow is disabled, we use 3-channel frame-to-frame RGB differences. Finally, we concatenate channels to form a 6-channel tensor per frame,
\begin{equation} 
Z_t=\begin{cases}
\big[u_t, v_t, \sqrt{u_t^2+v_t^2}\big] & \text{\small(3-ch flow)},\\
\big[U_t \,\Vert\, X^{\mathrm{ROI}}_t\big] & \text{\small(6-ch motion\,$\Vert$\,appear.)},\\
\big[\Delta X^{\mathrm{ROI}}_t \,\Vert\, X^{\mathrm{ROI}}_t\big] & \text{\small(no-flow ablation)}.
\end{cases}
\label{eq:optical-flow-channels}
\end{equation}
and uniformly chunk $Z_{1:T}$ for training or validation.

Body ROI uses a robust, rectangular crop centered on person detection; chest ROI is a square, face-aware crop anchored to the torso and gently biased toward the head. Both are temporally stabilized via median centers and upper-quantile sizes. This design enables a fair direct comparison of ROI choices under identical flow and model settings.

\subsection{Spatiotemporal Neural Network}
\label{sec:spatiotemporal}

The cropped optical flow stream is passed into a suite of spatiotemporal neural networks designed for physiological signal processing. We experiment with a suite of models, including DeepPhys \cite{chen2018deepphys}, MTTS-CAN \cite{liu2020multi}, EfficientPhys \cite{liu2023efficientphys}, and the model by Manne et al. designed for infant respiration, AIRFlowNet \cite{manne_automatic_2023}. Following the original dual-stream design, AIRFlowNet and EfficientPhys use 3-channel flow tensor comprising $u$, $v$, and magnitude $m=\sqrt{u^2+v^2}$. DeepPhys and MTTS-CAN receive a 6-channel tensor per frame composed of both motion branch (flow or RGB-difference) and appearance branch (raw frames). 

From the cropped clip $X^{\mathrm{ROI}}_{1:T}$ and (optionally) its dense flow $U_{1:T}$, we construct framewise inputs $Z_t \in \mathbb{R}^{h\times w\times C}$ (defined in \Cref{sec:preprocessing}). Models output a framewise waveform $\hat{y}_{1:T}$ that approximates the ground truth respiratory waveform $y_{1:T}$. Because infant recordings exhibit heterogeneous durations and frame rates, training and validation are computed at the \emph{subject} level. Mini-batch chunks are re-assembled per subject $s$ to form contiguous signals $\hat{y}^{(s)}_{1:T_s}$ and $y^{(s)}_{1:T_s}$ before the loss is applied. This stabilizes optimization in the presence of variable-length clips and yields unbiased validation across subjects.

\textbf{Loss.} We adopt a frequency-domain objective that compares \emph{normalized} power spectral densities (PSD) of prediction and target within a physiologically plausible band. For zero-mean signals, let $P_{\hat{y}}$ and $P_y$ denote their PSDs (via real fast Fourier transform). 
With $\tilde{P}(k)=\displaystyle \frac{P(k)}{\sum_{j\in\mathcal{K}} P(j)+\varepsilon}$ over bins $\mathcal{K}$ inside the band (with small $\varepsilon>0$ for numerical stability), the loss is
\begin{equation} 
\mathcal{L}_\text{PSD}
=\frac{1}{|\mathcal{K}|}\sum_{k\in\mathcal{K}}\!\big(\tilde{P}_{\hat{y}}(k)-\tilde{P}_{y}(k)\big)^2.
\label{eq:loss-psd}
\end{equation}
We average $\mathcal{L}_\text{PSD}$ over subjects in the batch. 
This emphasizes correct spectral concentration around the respiratory fundamental while being robust to small shifts in phase and amplitude.

\subsection{Post-Processing}

At inference we aggregate all chunks per subject and apply a standardized pipeline to obtain BPM. We first undo differencing if applied in preprocessing, and then remove low-frequency drift with a second-order smoother: 
\begin{equation} 
\hat{x}=\big(I+( \lambda^2 D^\top D)\big)^{-1}x,\quad x_\text{detrend}=x-\hat{x}
\label{eq:smoother},
\end{equation}
where $D$ is the second-order difference operator and $\lambda{=}100$ by default. We apply a first-order Butterworth band-pass with subject-specific bounds (0.3--1.0\,Hz, or 18--60\,BPM). Filtering uses zero-phase \texttt{filtfilt} to avoid lag. Finally, we estimate the rate by detecting peaks of the filtered waveform, computing the mean inter-peak interval $\Delta t$, and returning $60/\Delta t$ BPM.

\section{Evaluation and Results}
\label{sec:results}

We provide a comprehensive evaluation of various elements of the infant respiration estimation pipeline, including the choice of optical flow method, spatiotemporal network, ROI detection method, and the effect of training set size. We conducted experiments on our AIR-400 dataset, with 400 videos from 18 infant subjects. Spatiotemporal network and training configurations were standard and are detailed in the Supplementary Material. To robustly assess model performance, we performed all tests under six-fold, subject-wise cross-validation. In each fold, subjects were split randomly into training (12 subjects), validation (3), and testing (3), ensuring disjoint sets. All reported metrics are averaged across the six folds. Evaluation metrics included mean absolute error (MAE), root mean squared error (RMSE), and Pearson correlation ($\rho$) between predicted and ground truth respiration signals.

\subsection{Comprehensive Evaluation}

\label{sec:results-comprehensive}

\begin{table*}[!ht]
\centering
\caption{\textbf{Comprehensive evaluation of optical flow and spatiotemporal model pipelines,} under six-fold subject-wise cross-validation on AIR-400, with the infant body ROI detector enabled. Performance measured via mean absolute error (MAE), root mean squared error (RMSE), and Pearson correlation ($\rho$), each averaged across the six folds. Metrics are computed on post-processed respiration rates and averaged per subject within each fold. Best overall results by model in \textbf{bold}, and by optical flow (OF) method in \textit{italics}. Across models, classical optical flow methods consistently outperformed RAFT.}
\vspace{-2mm}
\label{tab:cv_results}
\setlength{\tabcolsep}{6pt}
\resizebox{\linewidth}{!}{
\begin{tabular}{lcccccccccccc}
\toprule
  \diagbox{\textbf{\textsc{OF}}}{\textbf{\textsc{Model}}}&  \multicolumn{3}{c}{DeepPhys \cite{chen2018deepphys}} & \multicolumn{3}{c}{MTTS-CAN \cite{liu2020multi}} & \multicolumn{3}{c}{EfficientPhys \cite{liu2023efficientphys}} & \multicolumn{3}{c}{AIRFlowNet \cite{manne_automatic_2023}} \\
\cmidrule(lr){2-4}\cmidrule(lr){5-7}\cmidrule(lr){8-10}\cmidrule(lr){11-13}
 & \textbf{\textsc{MAE ↓}} & \textbf{\textsc{RMSE ↓}} & \textbf{\textsc{$\rho$ ↑}} & \textbf{\textsc{MAE ↓}} & \textbf{\textsc{RMSE ↓}} & \textbf{\textsc{$\rho$ ↑}} & \textbf{\textsc{MAE ↓}} & \textbf{\textsc{RMSE ↓}} & \textbf{\textsc{$\rho$ ↑}} & \textbf{\textsc{MAE ↓}} & \textbf{\textsc{RMSE ↓}} & \textbf{\textsc{$\rho$ ↑}}\\
\midrule
Coarse2Fine \cite{liu2009beyond} &  4.26 & 6.82 & 0.45 & 5.16 & 8.30 & 0.28 & \textit{\textbf{3.70}} &\textit{\textbf{ 5.94}} &\textit{ \textbf{0.58}} & \textbf{3.84} & \textbf{6.67} & \textbf{0.44} \\
DeepFlow \cite{weinzaepfel2013deepflow} &  \textit{\textbf{3.88} }&\textit{ \textbf{6.45} }&\textit{\textbf{ 0.51} }& 5.05 & 8.21 & 0.36 & 5.01 & 7.54 & 0.34 & 4.41 & 7.56 & 0.39 \\
Farneb\"ack \cite{farneback2003two} &  4.97 & 8.28 & 0.27 & 4.56 & 7.72 & 0.44 & \textit{4.22} & \textit{6.84} & \textit{0.44} & 4.91 & 8.00 & 0.38 \\
PCAFlow \cite{wulff2015efficient} &  5.04 & 7.69 & 0.41 & \textit{\textbf{4.23}} & \textit{\textbf{6.85}} & \textit{\textbf{0.50}} & 4.55 & 7.11 & 0.42 & 4.25 & 6.95 & 0.47  \\
TV-L1 \cite{zach2007aduality,sanchez2013tvl1} &  4.48 & 7.11 & 0.39 & 4.91 & 7.97 & 0.42 & \textit{4.14} & \textit{6.59} & \textit{0.47} & 5.01 & 8.17 & 0.44  \\
RAFT \cite{zachary2020raft} &  \textit{6.47} & \textit{8.36} & \textit{0.35} & 23.07 & 26.23 & -0.17 & 12.77 & 14.57 & 0.10 & 7.52 & 9.85 & 0.26  \\
\bottomrule
\end{tabular} 
}
\end{table*}

We start with a evaluation of pipelines, pairing the different optical flow methods and spatiotemporal neural networks described in \Cref{sec:methodology}. \Cref{tab:cv_results} aggregates these results, all under six-fold cross-validation on AIR-400. Since the relative performance of these pipelines is not strongly affected by the ROI method, we highlight the reliable body-ROI results here and return to ROI comparisons below. 

\begin{figure}[!ht]
    \centering
    \includegraphics[width=1\linewidth]{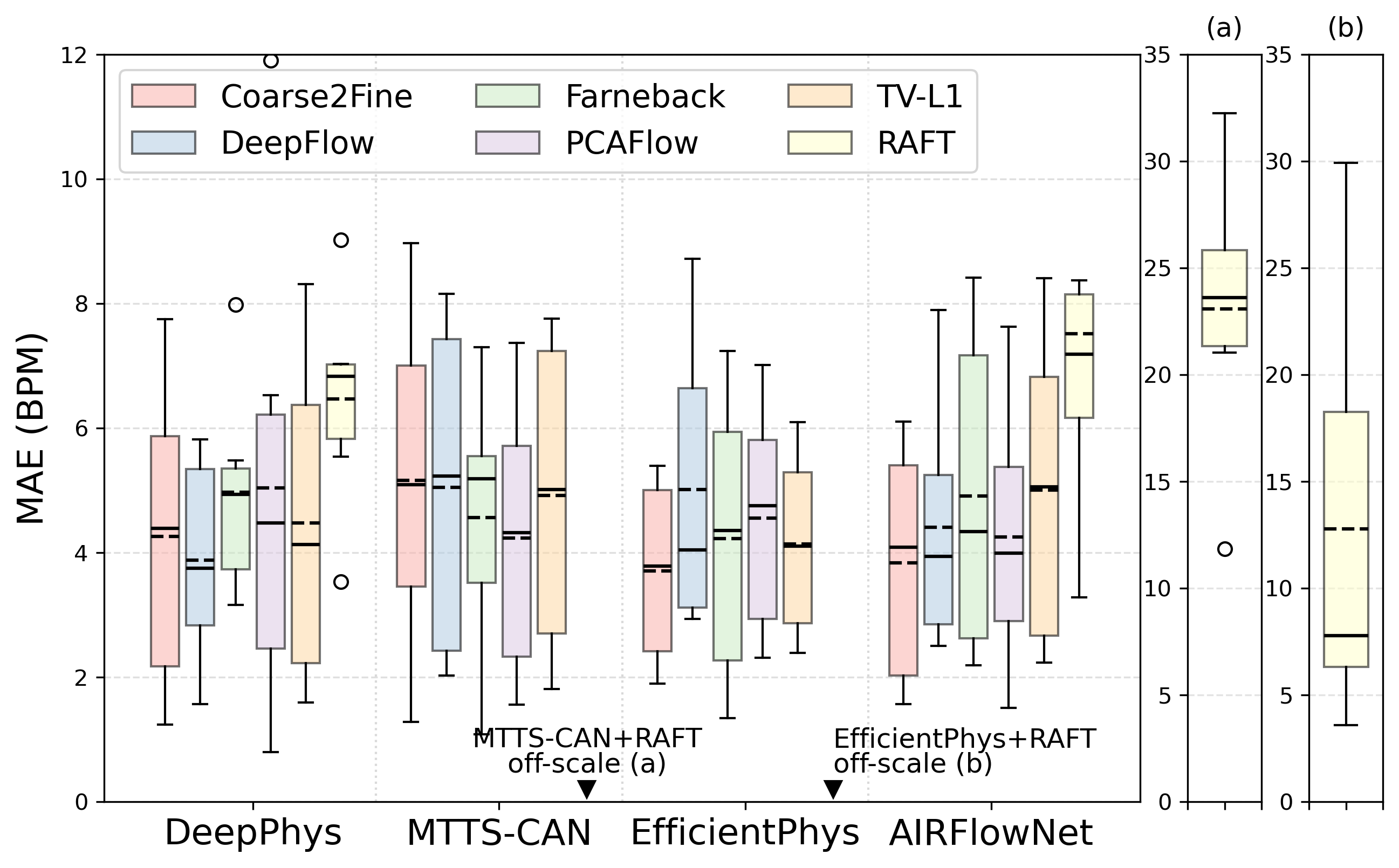}
    \vspace{-6mm}
    \caption{\textbf{Performance variation across cross-validation folds in our comprehensive evaluation} of infant respiration pipelines. Each box and whisker illustrates the MAE performance distribution in each of the six folds of AIR-400, with the box spanning the inter-quartile range (IQR) and whiskers extending to $1.5\times$IQR. The solid horizontal line inside each box denotes the median, and the dashed line denotes the mean.}
    \label{fig:cv_results}
\end{figure}

Our findings show that, outside of anomalous results under RAFT optical flow (discussed in the Supplementary Material), most models perform fairly capably, with many models achieving strong MAE scores near or below 4, and Pearson correlation ($\rho)$ over 0.5. This falls short of the performance of the best models on adult respiration estimation, which achieve MAEs in the 1--2 range and $\rho$s of 0.5--0.8 on the COHFACE dataset, but the adult subjects are stationary and training and testing is supported by more precise, sensor-based ground truth signals. Older spatiotemporal models such as DeepPhys and MTTS-CAN perform competitively with newer models, suggesting that the available data has not saturated the capabilities of the newer, more complex models. Despite the enlarged subject pool in AIR-400, the variation between folds in the cross validation is still relatively high, as shown in \Cref{fig:cv_results}. Thus, infant respiration estimation would likely be served by even larger datasets, especially in the absence of more precise ground truth respiratory signal capture.

\subsection{Effect of Training Set Size}

\label{sec:training-size}

\begin{figure}[!ht]
    \centering
    \includegraphics[width=\linewidth]{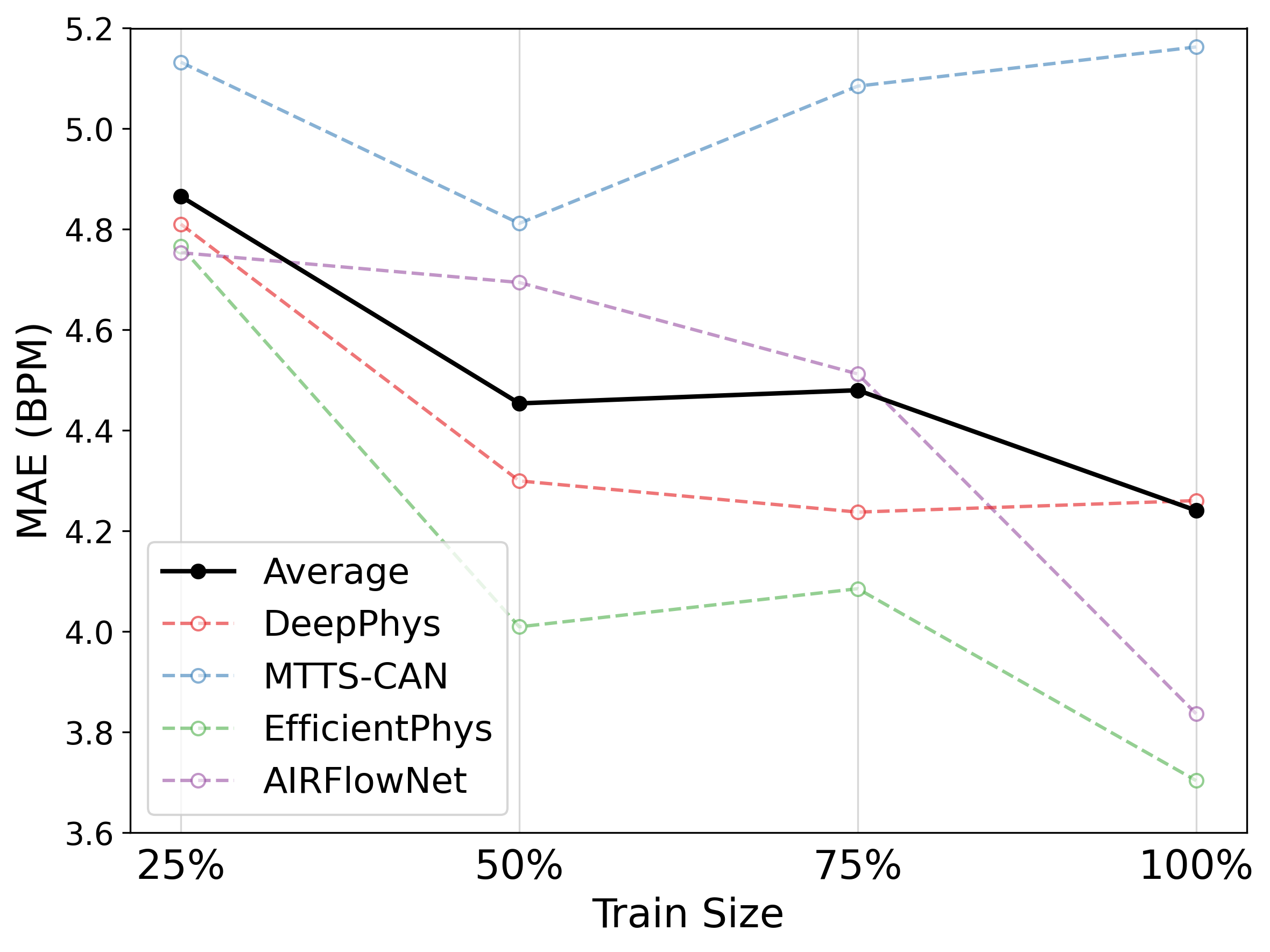}
    \vspace{-8mm}
    \caption{\textbf{Effect of training set size on respiration estimation.} Line charts show the performance of the DeepPhys, MTTS-CAN, EfficientPhys, and AIRFlowNet models---and their average---under 6-fold cross-validation, when training is restricted to random subsets of the indicated size. These results show, consistently across several models, the significant improvements enabled by the larger training set sizes from AIR-400.}
    \label{fig:train_size_results}
    \vspace{-4mm}
\end{figure}

We further explore the small data constraints on infant respiration estimation. \Cref{fig:train_size_results} shows the effect of restricting the training set size, to various proportions by subject, across our suite of models. There is a clear overall trend of greater training data availability leading to stronger results, but the small dataset sizes leads to stochasticity that contravenes this effect in certain instances. Nonetheless, the benefit of increasing the dataset size does not appear to have saturated in this complex task. We do not specifically compare to training with AIR-125 here, due to the complexity of integrating its unevenly represented subjects into cross-validation on AIR-400, but the evaluation at the 25\% training set size serves as an approximate substitute. 

\subsection{Comparison of ROI Methods}

\label{sec:roi}

\Cref{tab:roi-comparison} offers a performance comparison of our body ROI and chest ROI detection methods in the respiration estimation pipeline. We test under six-fold cross-validation across a range of models, with the optical flow method fixed. Overall, the body ROI performs better than no ROI for three of the four architectures, whereas the more concentrated chest ROI yields mixed results, sometimes slightly degrading performance. Qualitatively, we find that the body ROI method provides the most reliable and stable view of infant respiration activity across subjects, but additional work is needed to translate this into consistent gains. The Supplementary Material discusses challenges with the chest ROI detection.

\begin{table}[!ht]
\centering
\caption{\textbf{Performance evaluation of ROI methods} across different models, under six-fold cross-validation on AIR-400, with Coarse2Fine optical flow processing. Strongest overall results per model in \textbf{bold}. Using the infant body ROI generally improves MAE and RMSE for DeepPhys, EfficientPhys, and AIRFlowNet, but not MTTS-CAN.}
\vspace{-2mm}
\label{tab:roi-comparison}
\setlength{\tabcolsep}{6pt}
\footnotesize
\resizebox{0.85\linewidth}{!}{
\begin{tabular}{llrrr}
\toprule
\textbf{\textsc{Model}} & \textbf{\textsc{ROI}} & \textbf{\textsc{MAE ↓}} & \textbf{\textsc{RMSE ↓}} & \textbf{\textsc{$\rho$ ↑}} \\
\midrule
\multirow{3}{5em}{DeepPhys}  & None &  4.78 & 7.69 & 0.33 \\
  & \textbf{Body ROI} & \textbf{4.26} & \textbf{6.82} & \textbf{0.45} \\
  & Chest ROI & 4.45 & 6.92 & 0.39 \\
\midrule
\multirow{3}{5em}{MTTS-CAN} & \textbf{None} & \textbf{4.35 }& \textbf{7.59 }& \textbf{0.38} \\
 & Body ROI & 5.16 & 8.30 & 0.28 \\
 & Chest ROI & 4.90 & 7.67 & 0.38\\
\midrule
\multirow{3}{5em}{EfficientPhys}  & None & 4.57& 7.08 & 0.39 \\
  & \textbf{Body ROI} & \textbf{3.70} & \textbf{5.94} & \textbf{0.58} \\
  & Chest ROI & 4.68 & 7.23 & 0.38\\
\midrule
\multirow{3}{5em}{AIRFlowNet}   & None & 4.76 & 7.79 & 0.32 \\
 & \textbf{Body ROI} & \textbf{3.84} & \textbf{6.67} & \textbf{0.44} \\
  & Chest ROI & 4.24 & 7.14 & 0.42\\
\bottomrule
\end{tabular}
}
\vspace{-0.5em}
\end{table}

\subsection{Experimental Reproducibility and Variability}

\label{sec:reproducibility}

\begin{figure}[!ht]
    \centering
    \includegraphics[width=\linewidth]{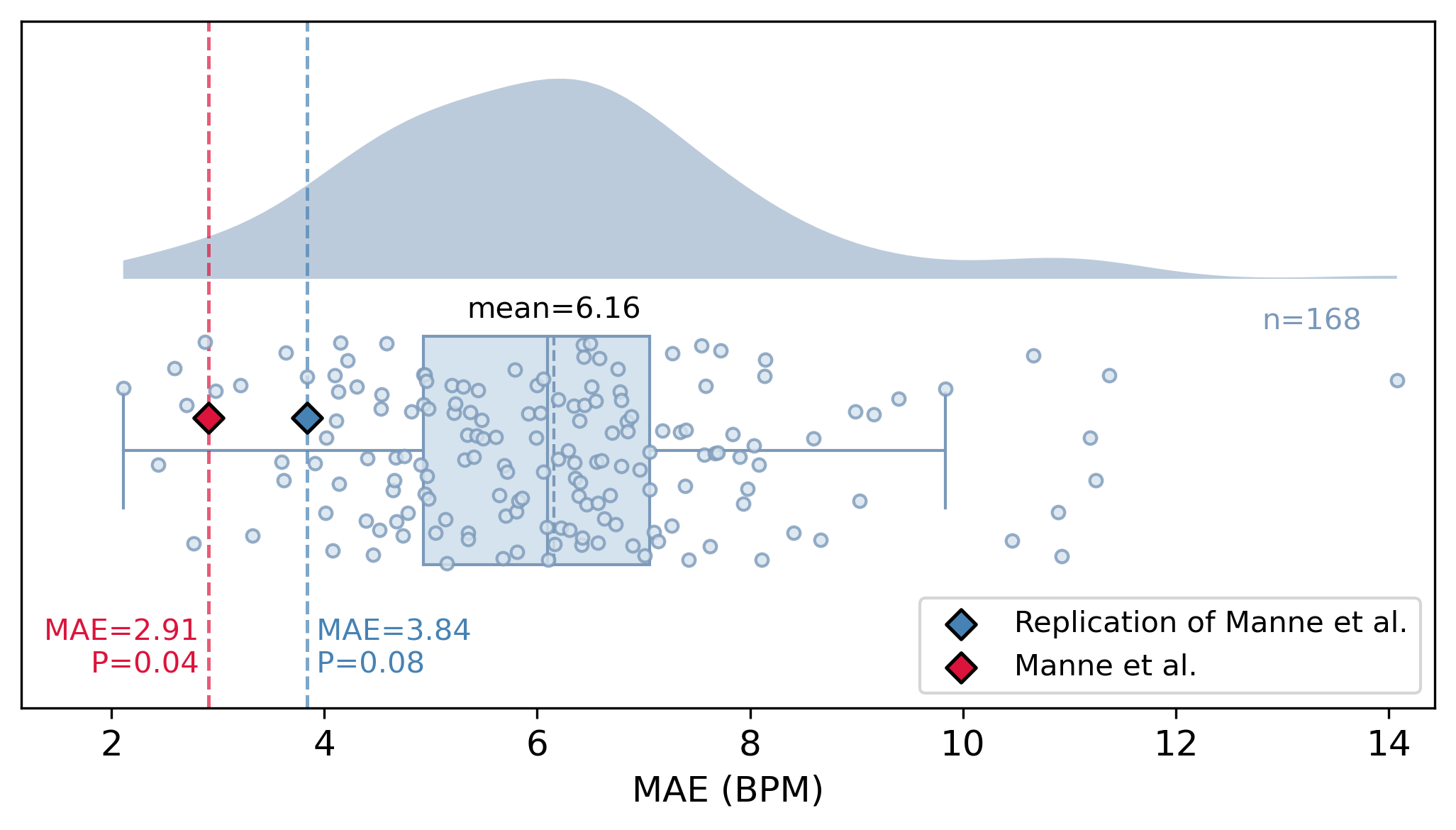}
    \vspace{-8mm}
    \caption{\textbf{Reproducing Manne et al.} Illustrating how reported and recreated results from \cite{manne_automatic_2023} sit as outliers in the distribution of MAE results under all 168 train--test configurations in AIR-125 consistent with their documented specifications. We argue that our present, comprehensive cross-validation outcomes on AIR-400 represent more reproducible baselines for infant respiration estimation. The box plot shows the mean, median, and interquartile range (IQR) with whiskers at $1.5\times\text{IQR}$; the half-violin plot above shows a corresponding kernel density estimation.}
    \label{fig:air125_combos}
    \vspace{-2mm}
\end{figure}

In \Cref{sec:results-comprehensive}, we established under comprehensive cross-validation that our best infant respiration models perform with MAEs of 3.7--4.0\,BPM, and in \Cref{sec:training-size}, we showed that the larger training sets afforded by AIR-400 are essential for achieving this performance. These findings stand in apparent tension with results from our earlier infant respiration efforts in Manne et al. \cite{manne_automatic_2023}, where our best model, AIRFlowNet, achieved a MAE of 2.91\,BPM, despite being trained on the smaller but similar AIR-125 dataset. We conducted a careful study to resolve this discrepancy, described in the Supplementary Materials. Here, we note our main findings: that (i) the precise results were not easily reproducible \cite{manne_automatic_2023}, but (ii) any results in the MAE 2.9--4.0\,BPM range sit well below the mean of 6.16\,BPM, corresponding to advantageous choices of subjects for the train--test split, and do not represent reliably achievable performance on AIR-125 (see \Cref{fig:air125_combos}).
\section{Conclusion \& Outlook}
\label{sec:conclusion}

In this work, we establish infant respiration estimation as a reproducible and benchmarkable task in computer vision for the first time, with the introduction of our AIR-400 dataset and our spatiotemporal estimation pipeline, as well as our comprehensive evaluation and resolution of reproducibility issues. Our results suggest that larger training datasets would support further performance gains, including perhaps by enabling estimation pipelines to make better use of region-of-interest detection and greater model capacities. Our study focuses on hand-curated clips featuring mostly-still infants with distinct breathing patterns, leaving room for future approaches to tackle more active subjects and noisier footage. In addition, downstream applications to the detection of breathing irregularities such as apnea or respiratory distress would be broadly impactful.

\clearpage
\section*{Acknowledgments}
\label{sec:acknowledgements}

We gratefully acknowledge support for this research from a Northeastern University TIER 1 Seed Grant, and NSF CAREER Award \#2143882.

{
    \small
    \bibliographystyle{ieeenat_fullname}
    \bibliography{refs}

@String(CVPR= {IEEE Conf. Comput. Vis. Pattern Recog.})

@String(ECCV= {Eur. Conf. Comput. Vis.})

@String(ICPR = {Int. Conf. Pattern Recog.})

@String(ICIP = {IEEE Int. Conf. Image Process.})

@String(CVPRW= {IEEE Conf. Comput. Vis. Pattern Recog. Worksh.})

@String(CVPR  = {CVPR})

@String(ECCV  = {ECCV})

@String(ICPR  = {ICPR})

@String(ICIP  = {ICIP})

@String(CVPRW= {CVPRW})

@article{ostadabbas2015vision,
  title={A vision-based respiration monitoring system for passive airway resistance estimation},
  author={Ostadabbas, Sarah and Sebkhi, Nordine and Zhang, Mingxi and Rahim, Salman and Anderson, Larry J and Lee, Frances Eun-Hyung and Ghovanloo, Maysam},
  journal={IEEE Transactions on biomedical engineering},
  volume={63},
  number={9},
  pages={1904--1913},
  year={2015},
  publisher={IEEE}
}

@inproceedings{ostadabbas2014passive,
  title={A passive quantitative measurement of airway resistance using depth data},
  author={Ostadabbas, Sarah and Bulach, Christoph and Ku, David N and Anderson, Larry J and Ghovanloo, Maysam},
  booktitle={2014 36th Annual International Conference of the IEEE Engineering in Medicine and Biology Society},
  pages={5743--5747},
  year={2014},
  organization={IEEE}
}

@ARTICLE{dechemi_ebabynet_2024,
  author={Dechemi, Amel and Karydis, Konstantinos},
  journal={IEEE Transactions on Neural Systems and Rehabilitation Engineering}, 
  title={E-BabyNet: Enhanced Action Recognition of Infant Reaching in Unconstrained Environments}, 
  year={2024},
  volume={32},
  number={},
  pages={1679-1686},
  doi={10.1109/TNSRE.2024.3392161}}

@misc{yu_remote_2019,
	title = {Remote {Photoplethysmograph} {Signal} {Measurement} from {Facial} {Videos} {Using} {Spatio}-{Temporal} {Networks}},
	url = {http://arxiv.org/abs/1905.02419},
	doi = {10.48550/arXiv.1905.02419},
	language = {en},
	urldate = {2025-09-17},
	publisher = {arXiv},
	author = {Yu, Zitong and Li, Xiaobai and Zhao, Guoying},
	month = jul,
	year = {2019},
	note = {arXiv:1905.02419 [cs]},
}

@phdthesis{liu2009beyond,
  title        = {Beyond Pixels: Exploring New Representations and Applications for Motion Analysis},
  author       = {Liu, Ce},
  school       = {Massachusetts Institute of Technology},
  year         = {2009},
  month        = {May},
  type         = {Doctoral Thesis},
  address      = {Cambridge, MA, USA}
}

@INPROCEEDINGS{weinzaepfel2013deepflow,
  author={Weinzaepfel, Philippe and Revaud, Jerome and Harchaoui, Zaid and Schmid, Cordelia},
  booktitle={2013 IEEE International Conference on Computer Vision}, 
  title={DeepFlow: Large Displacement Optical Flow with Deep Matching}, 
  year={2013},
  volume={},
  number={},
  pages={1385-1392},
  doi={10.1109/ICCV.2013.175}}

@inproceedings{farneback2003two,
  author       = {Farneb{\"a}ck, Gunnar},
  title        = {Two-Frame Motion Estimation Based on Polynomial Expansion},
  booktitle    = {Image Analysis (SCIA 2003)},
  editor       = {Bigun, Josef and Gustavsson, Tomas},
  series       = {Lecture Notes in Computer Science},
  volume       = {2749},
  pages        = {363--370},
  year         = {2003},
  publisher    = {Springer}
}

@inproceedings{wulff2015efficient,
  author={Wulff, Jonas and Black, Michael J.},
  booktitle={2015 IEEE Conference on Computer Vision and Pattern Recognition (CVPR)}, 
  title={Efficient sparse-to-dense optical flow estimation using a learned basis and layers}, 
  year={2015},
  volume={},
  number={},
  pages={120-130},
  doi={10.1109/CVPR.2015.7298607}}

@INPROCEEDINGS{ali2014robust,
  author={Ali, Sharib and Daul, Christian and Blondel, Walter},
  booktitle={2014 4th International Conference on Image Processing Theory, Tools and Applications (IPTA)}, 
  title={Robust and accurate optical flow estimation for weak texture and varying illumination conditions: Application to cystoscopy}, 
  year={2014},
  volume={},
  number={},
  pages={1-6},
  doi={10.1109/IPTA.2014.7001947}}

@inproceedings{zach2007aduality,
  author = {Zach, Christopher and Pock, Thomas and Bischof, Horst},
  booktitle = {DAGM-Symposium},
  editor = {Hamprecht, Fred A. and Schnörr, Christoph and Jähne, Bernd},
  ee = {https://doi.org/10.1007/978-3-540-74936-3_22},
  isbn = {978-3-540-74933-2},
  pages = {214-223},
  publisher = {Springer},
  series = {Lecture Notes in Computer Science},
  title = {A Duality Based Approach for Realtime TV-L1 Optical Flow.},
  volume = {4713},
  year = {2007}
}

@article{sanchez2013tvl1,
    title   = {{TV-L1 Optical Flow Estimation}},
    author  = {Sánchez Pérez, Javier and Meinhardt-Llopis, Enric and Facciolo, Gabriele},
    journal = {{Image Processing On Line}},
    volume  = {3},
    pages   = {137--150},
    year    = {2013},
    note    = {\url{https://doi.org/10.5201/ipol.2013.26}}
}

@misc{zachary2020raft,
      title={RAFT: Recurrent All-Pairs Field Transforms for Optical Flow}, 
      author={Zachary Teed and Jia Deng},
      year={2020},
      eprint={2003.12039},
      archivePrefix={arXiv},
      primaryClass={cs.CV},
      url={https://arxiv.org/abs/2003.12039}, 
}

@software{Jocher_Ultralytics_YOLO_2023,
author = {Jocher, Glenn and Qiu, Jing and Chaurasia, Ayush},
license = {AGPL-3.0},
month = jan,
title = {{Ultralytics YOLO}},
url = {https://github.com/ultralytics/ultralytics},
version = {8.0.0},
year = {2023}
}

@inproceedings{wan_infanface_2022,
	address = {Montreal, QC, Canada},
	title = {{InfAnFace}: Bridging the Infant-Adult Domain Gap in Facial Landmark Estimation in the Wild},
	isbn = {978-1-6654-9062-7},
	shorttitle = {{InfAnFace}},
	url = {https://ieeexplore.ieee.org/document/9956647/},
	doi = {10.1109/ICPR56361.2022.9956647},
	language = {en},
	urldate = {2022-12-02},
	booktitle = {2022 26th {International} {Conference} on {Pattern} {Recognition} ({ICPR})},
	publisher = {IEEE},
	author = {Wan, Michael and Zhu, Shaotong and Luan, Lingfei and Prateek, Gulati and Huang, Xiaofei and Schwartz-Mette, Rebecca and Hayes, Marie and Zimmerman, Emily and Ostadabbas, Sarah},
	month = aug,
	year = {2022},
	pages = {4486--4492},
}

@incollection{manne_automatic_2023,
	address = {Cham},
	title = {Automatic {Infant} {Respiration} {Estimation} from {Video}: {A} {Deep} {Flow}-{Based} {Algorithm} and a {Novel} {Public} {Benchmark}},
	volume = {14246},
	isbn = {978-3-031-45543-8 978-3-031-45544-5},
	shorttitle = {Automatic {Infant} {Respiration} {Estimation} from {Video}},
	url = {https://link.springer.com/10.1007/978-3-031-45544-5_10},
	language = {en},
	urldate = {2023-10-23},
	booktitle = {Perinatal, {Preterm} and {Paediatric} {Image} {Analysis}},
	publisher = {Springer Nature Switzerland},
	author = {Manne, Sai Kumar Reddy and Zhu, Shaotong and Ostadabbas, Sarah and Wan, Michael},
	editor = {Link-Sourani, Daphna and Abaci Turk, Esra and Macgowan, Christopher and Hutter, Jana and Melbourne, Andrew and Licandro, Roxane},
	year = {2023},
	doi = {10.1007/978-3-031-45544-5_10},
	note = {Series Title: Lecture Notes in Computer Science},
	pages = {111--120},
}

@inproceedings{huang_posture-based_2023,
	address = {Vancouver, BC, Canada},
	title = {Posture-based {Infant} {Action} {Recognition} in the {Wild} with {Very} {Limited} {Data}},
	copyright = {https://doi.org/10.15223/policy-029},
	isbn = {979-8-3503-0249-3},
	url = {https://ieeexplore.ieee.org/document/10208410/},
	doi = {10.1109/CVPRW59228.2023.00519},
	language = {en},
	urldate = {2024-10-02},
	booktitle = {2023 {IEEE}/{CVF} {Conference} on {Computer} {Vision} and {Pattern} {Recognition} {Workshops} ({CVPRW})},
	publisher = {IEEE},
	author = {Huang, Xiaofei and Luan, Lingfei and Hatamimajoumerd, Elaheh and Wan, Michael and Kakhaki, Pooria Daneshvar and Obeid, Rita and Ostadabbas, Sarah},
	month = jun,
	year = {2023},
	pages = {4912--4921},
}

@inproceedings{tveit2016motion,
  title={Motion based detection of respiration rate in infants using video},
  author={Tveit, Daniel Myklatun and Engan, Kjersti and Austvoll, Ivar and Meinich-Bache, {\O}yvind},
  booktitle={2016 IEEE International Conference on Image Processing (ICIP)},
  pages={1225--1229},
  year={2016},
  organization={IEEE}
}

@inproceedings{chen2018deepphys,
  title={Deepphys: Video-based physiological measurement using convolutional attention networks},
  author={Chen, Weixuan and McDuff, Daniel},
  booktitle={Proceedings of the european conference on computer vision (ECCV)},
  pages={349--365},
  year={2018}
}

@article{liu2020multi,
  title={Multi-task temporal shift attention networks for on-device contactless vitals measurement},
  author={Liu, Xin and Fromm, Josh and Patel, Shwetak and McDuff, Daniel},
  journal={Advances in Neural Information Processing Systems},
  volume={33},
  pages={19400--19411},
  year={2020}
}

@article{lorato2021towards,
  title={Towards continuous camera-based respiration monitoring in infants},
  author={Lorato, Ilde and Stuijk, Sander and Meftah, Mohammed and Kommers, Deedee and Andriessen, Peter and van Pul, Carola and de Haan, Gerard},
  journal={Sensors},
  volume={21},
  number={7},
  pages={2268},
  year={2021},
  publisher={MDPI}
}

@inproceedings{liu2023efficientphys,
  title={EfficientPhys: Enabling Simple, Fast and Accurate Camera-Based Cardiac Measurement},
  author={Liu, Xin and Hill, Brian and Jiang, Ziheng and Patel, Shwetak and McDuff, Daniel},
  booktitle={Proceedings of the IEEE/CVF Winter Conference on Applications of Computer Vision},
  pages={5008--5017},
  year={2023}
}

@article{foldesy2020reference,
  title={Reference free incremental deep learning model applied for camera-based respiration monitoring},
  author={F{\"o}ldesy, P{\'e}ter and Zar{\'a}ndy, {\'A}kos and Szab{\'o}, Mikl{\'o}s},
  journal={IEEE Sensors Journal},
  volume={21},
  number={2},
  pages={2346--2352},
  year={2020},
  publisher={IEEE}
}

@inproceedings{li2018obf,
  title={The obf database: A large face video database for remote physiological signal measurement and atrial fibrillation detection},
  author={Li, Xiaobai and Alikhani, Iman and Shi, Jingang and Seppanen, Tapio and Junttila, Juhani and Majamaa-Voltti, Kirsi and Tulppo, Mikko and Zhao, Guoying},
  //booktitle={2018 13th IEEE international conference on automatic face \& gesture recognition (FG 2018)},
  booktitle={2018 13th IEEE international conference on automatic face \& gesture recognition (FG 2018)},
  pages={242--249},
  year={2018},
  organization={IEEE}
}

@article{reuter2014respiratory,
    author = {Reuter, Suzanne and Moser, Chuanpit and Baack, Michelle},
    title = "{Respiratory Distress in the Newborn}",
    journal = {Pediatrics In Review},
    volume = {35},
    number = {10},
    pages = {417-429},
    year = {2014},
    month = {10},
    issn = {0191-9601}
}

@article{villarroel_non-contact_2019,
	title = {Non-contact physiological monitoring of preterm infants in the {Neonatal} {Intensive} {Care} {Unit}},
	volume = {2},
	issn = {2398-6352},
	language = {en},
	number = {1},
	urldate = {2023-06-26},
	journal = {npj Digital Medicine},
	author = {Villarroel, Mauricio and Chaichulee, Sitthichok and Jorge, João and Davis, Sara and Green, Gabrielle and Arteta, Carlos and Zisserman, Andrew and McCormick, Kenny and Watkinson, Peter and Tarassenko, Lionel},
	month = dec,
	year = {2019},
	pages = {128},
}

@inproceedings{kyrollos2021noncontact,
  title={Noncontact neonatal respiration rate estimation using machine vision},
  author={Kyrollos, Daniel G and Tanner, Joshua B and Greenwood, Kim and Harrold, JoAnn and Green, James R},
  booktitle={2021 IEEE Sensors Applications Symposium (SAS)},
  pages={1--6},
  year={2021},
  organization={IEEE}
}

@inproceedings{dutta2019vgg,
  author = {Dutta, Abhishek and Zisserman, Andrew},
  title = {The {VIA} Annotation Software for Images, Audio and Video},
  booktitle = {Proceedings of the 27th ACM International Conference on Multimedia},
  series = {MM '19},
  year = {2019},
  isbn = {978-1-4503-6889-6/19/10},
  location = {Nice, France},
  numpages = {4},
  url = {https://doi.org/10.1145/3343031.3350535},
  doi = {10.1145/3343031.3350535},
  publisher = {ACM},
  address = {New York, NY, USA},
}

@article{mcduff2022scamps,
  title={Scamps: Synthetics for camera measurement of physiological signals},
  author={McDuff, Daniel and Wander, Miah and Liu, Xin and Hill, Brian and Hernandez, Javier and Lester, Jonathan and Baltrusaitis, Tadas},
  journal={Advances in Neural Information Processing Systems},
  volume={35},
  pages={3744--3757},
  year={2022}
}

@article{heusch2017reproducible,
  title={A reproducible study on remote heart rate measurement},
  author={Heusch, Guillaume and Anjos, Andr{\'e} and Marcel, S{\'e}bastien},
  journal={arXiv preprint arXiv:1709.00962},
  year={2017}
}

@article{soleymani2011multimodal,
  title={A multimodal database for affect recognition and implicit tagging},
  author={Soleymani, Mohammad and Lichtenauer, Jeroen and Pun, Thierry and Pantic, Maja},
  journal={IEEE transactions on affective computing},
  volume={3},
  number={1},
  pages={42--55},
  year={2011},
  publisher={IEEE}
}

@inproceedings{estepp2014recovering,
  title={Recovering pulse rate during motion artifact with a multi-imager array for non-contact imaging photoplethysmography},
  author={Estepp, Justin R and Blackford, Ethan B and Meier, Christopher M},
  booktitle={2014 IEEE international conference on systems, man, and cybernetics (SMC)},
  pages={1462--1469},
  year={2014},
  organization={IEEE}
}

@inproceedings{huang2021invariant,
  title={Invariant representation learning for infant pose estimation with small data},
  author={Huang, Xiaofei and Fu, Nihang and Liu, Shuangjun and Ostadabbas, Sarah},
  booktitle={2021 16th IEEE International Conference on Automatic Face and Gesture Recognition (FG 2021)},
  pages={1--8},
  year={2021},
  organization={IEEE}
}

@article{janvier2004apnea,
  title={Apnea is associated with neurodevelopmental impairment in very low birth weight infants},
  author={Janvier, Annie and Khairy, May and Kokkotis, Athanasios and Cormier, Carole and Messmer, Denise and Barrington, Keith J},
  journal={Journal of perinatology},
  volume={24},
  number={12},
  pages={763--768},
  year={2004},
  publisher={Nature Publishing Group}
}

@article{bonner2017there,
  title={‘There were more wires than him’: the potential for wireless patient monitoring in neonatal intensive care},
  author={Bonner, Oliver and Beardsall, Kathryn and Crilly, Nathan and Lasenby, Joan},
  journal={BMJ innovations},
  volume={3},
  number={1},
  year={2017},
  publisher={BMJ Specialist Journals}
}

@article{eichenwald2016apnea,
  title={Apnea of prematurity},
  author={Eichenwald, Eric C and Watterberg, Kristi L and Aucott, Susan and Benitz, William E and Cummings, James J and Goldsmith, Jay and Poindexter, Brenda B and Puopolo, Karen and Stewart, Dan L and Wang, Kasper S and others},
  journal={Pediatrics},
  volume={137},
  number={1},
  year={2016},
  publisher={American Academy of Pediatrics}
}

@article{martin2022apnea,
  title={Apnea of prematurity and sudden infant death syndrome},
  author={Martin, Richard J and Mitchell, Lisa J and MacFarlane, Peter M},
  journal={Handbook of clinical neurology},
  volume={189},
  pages={43--52},
  year={2022},
  publisher={Elsevier}
}

@article{fleming_normal_2011,
    title={Normal ranges of heart rate and respiratory rate in children from birth to 18 years of age: a systematic review of observational studies},
    author={Fleming, Sina and Thompson, Matthew and Stevens, Richard and Heneghan, Carl and Pl{\"u}ddemann, Annette and Maconochie, Ian and Tarassenko, Lionel and Mant, David},
    journal={The Lancet},
    volume={377},
    number={9770},
    pages={1011--1018},
    year={2011},
    publisher={Elsevier},
    doi={10.1016/S0140-6736(10)62226-X}
}

@article{massaroni2024advances,
  author = {Vitazkova, Diana and Foltan, Erik and Kosnacova, Helena and Micjan, Michal and Donoval, Martin and Kuzma, Anton and Kopani, Martin and Vavrinsky, Erik},
  title = {Advances in Respiratory Monitoring: A Comprehensive Review of Wearable and Remote Technologies},
  journal={Biosensors},
  volume={14},
  year={2024},
  number={2},
  pages={90},
  doi={10.3390/bios14020090}
}

@article{santos2023realtime,
  title = {Real-time changes in rib cage expansion and use of abdominal mechanical stimulation in newborns: a quasi-experimental study},
  author = {Santos, J. X. and Silva, P. Y. F. and Cruz, M. C. L. D. and Silva, B. F. V. E. and Azevedo, I. G. and Pereira, S. A.},
  journal = {Revista Paulista de Pediatria},
  volume = {42},
  pages = {e2023032},
  year = {2023},
  doi = {10.1590/1984-0462/2024/42/2023032}
}

@article{kohn2015monitoring,
  title = {Monitoring the respiratory rate by miniature motion sensors in premature infants: a comparative study},
  author = {Kohn, S. and Waisman, D. and Pesin, J. and Huna-Baron, G. and Faingersh, A. and Landesberg, A. and Rotschild, A.},
  journal = {Journal of Perinatology},
  volume = {36},
  number = {4},
  pages = {289--294},
  year = {2015},
  doi = {10.1038/jp.2015.173}
}
}
\clearpage
\section*{Supplementary Material}
\label{sec:supplementary}

\subsection*{Supplement to \Cref{sec:results}: Training Configuration}

Unless otherwise noted, all spatiotemporal models are trained for 15 epochs with a fixed learning rate of $1\times 10^{-3}$ and a mini-batch size of 4. We fixed random seeds and enforced deterministic PyTorch settings for reproducibility. For validation, testing, and inference, we use a batch size of 8 to better utilize GPU memory. 

We did not employ early stopping; instead, for each experiment we select the checkpoint with the best validation performance according to the subject-averaged $\mathcal{L}_\text{PSD}$ in \eqref{eq:loss-psd} among the 15 epochs. In practice, the validation loss stabilizes well before the final epoch, and we did not observe divergence across runs. 

Our codebase also supports alternative training objectives, including negative Pearson correlation and time-domain mean squared error, but unless otherwise specified, all results in \Cref{sec:results} use the PSD-MSE objective. For reproducibility, the exact hyperparameters and loss configurations for each experiment are specified in the YAML configuration files included in our public code release.

\subsection*{Supplement to \Cref{sec:results-comprehensive}: RAFT Optical Flow Failure}
While classical optical flow methods (Coarse2Fine, DeepFlow, Farnebäck, PCAFlow, and TV-L1) yield consistent performance across models, we observe a dramatic degradation when using RAFT. Across all spatiotemporal networks, RAFT produces up to $5\times$ higher MAE and RMSE, and even negative correlations, indicating a systematic failure rather than stochastic variance. As visualized in Figure~\ref{fig:optical_flow_example}, RAFT predicts dense, large-magnitude motion across nearly the entire frame, including static background regions, instead of isolating subtle, periodic chest motion. 

We hypothesize that this behavior reflects a domain mismatch. RAFT was developed and evaluated primarily on large-displacement, high-texture motion benchmarks such as Sintel and KITTI~\cite{zachary2020raft}, whereas our infrared infant videos exhibit low texture, low signal-to-noise ratio (SNR), and only minute thoracoabdominal motion. In this setting, RAFT's dense all-pairs correlation volume may over-interpret small fluctuations and sensor noise as coherent motion, leading to physiologically implausible flow fields. 

By contrast, classical variational and coarse-to-fine methods such as TV-L1 explicitly regularize the flow with spatial smoothness terms and robust data penalties, and enforce multi-scale consistency through pyramid-based warping~\cite{zach2007aduality, sanchez2013tvl1}. These priors are known to suppress high-frequency noise and favor piecewise-smooth motion, and have been successfully applied to weak-texture, low-contrast medical imagery~\cite{ali2014robust}, consistent with our observation that they produce flows more localized to the thoracoabdominal region and support substantially lower respiration estimation errors.

\subsection*{Supplement to \Cref{sec:roi}: Chest ROI Inconsistency}

We additionally investigate why the more concentrated chest ROI does not consistently outperform the coarser body ROI. Our chest ROI is defined as a square inscribed within the body box (using its shorter side) and shifted slightly toward the head. As illustrated in \Cref{fig:roi_boxes}, this crop emphasizes the upper torso but truncates part of the lower abdomen, where supine infant breathing motion is often pronounced. Meanwhile, head, shoulder, and arm movements (as well as blankets or toys) are more prevalent in the upper region and therefore receive greater weight inside the chest ROI. 

By contrast, the body ROI preserves the full thoracoabdominal motion field while still restricting the input to the infant torso, giving the spatiotemporal networks sufficient context to internally attend to the most informative subregions. These factors provide a plausible explanation for why the body ROI yields more reliable improvements, whereas the chest ROI produces mixed results across architectures.

\subsection{Supplement to \Cref{sec:reproducibility}: Reproducibility in Manne et al. \cite{manne_automatic_2023}}

Regrettably, we could not recall or uncover documentation of the specific model and training configuration used in Manne et al., nor could we obtain precisely the same metric results with reconstructions, despite extensive efforts on both fronts. 

Our best attempt to reproduce the AIRFlowNet configuration used in \cite{manne_automatic_2023} yielded a model achieving a MAE of 3.84\,BPM, under what we believe to be the original train--test split, with three subjects chosen for training and five for testing. We then evaluated this replicate model under \emph{all} ${8\choose3}\times3=168$ train--test splits with three subjects chosen for training (one held for validation), and plotted the results in \Cref{fig:air125_combos}. These reveal a high variability in MAE by split choice (consistent with our findings on fold-variability in \Cref{fig:cv_results}), and a particularly low MAE of 3.84\,BPM achieved by the split likely used in Manne et al., compared to mean MAEs of 6.16\,BPM by split. We believe these higher MAEs in the 5.5--6.5\,BPM range more accurately reflect the true performance Manne et al.'s model in the AIR-125 dataset, and that our present results in the 3.7--4.0\,BPM range on AIR-400 reflect the current, generalizable state-of-the-art performance of infant respiration rate estimation from spatiotemporal models.

\end{document}